%% file: 0.main.tex
\documentclass{article}
\usepackage{ijcai25}

\usepackage{times}
\usepackage{soul}
\usepackage{url}
\usepackage[hidelinks]{hyperref}
\usepackage[utf8]{inputenc}
\usepackage[small]{caption}
\usepackage{graphicx}
\usepackage{amsmath}
\usepackage{amsthm}
\usepackage{booktabs}
\usepackage{algorithm}
\usepackage{algorithmic}
\usepackage[switch]{lineno}

\usepackage{pifont}
\usepackage{xcolor}
\newcommand{\cmark}{\textcolor{green!80!black}{\ding{51}}}
\newcommand{\xmark}{\textcolor{red}{\ding{55}}}
\usepackage{amsmath,scalerel}
\DeclareMathOperator*{\concat}{\scalerel*{\Vert}{\sum}}
\usepackage{svg}
\svgsetup{inkscapelatex=false}
\usepackage{amssymb}
\usepackage[utf8]{inputenc}
\usepackage[english]{babel}
\usepackage{hyperref} 
\hypersetup{ colorlinks=true, linkcolor=black, filecolor=black, urlcolor=cyan, }

\newcommand{\ccell}[2]{#1}

\title{Efficient Knowledge Tracing \\Leveraging Higher-Order Information in Integrated Graphs}

\author{
Donghee Han$^1$
\and
Daehee Kim$^1$
\and
Minjun Lee$^2$
\and
Daeyoung Roh$^1$
\and
Keejun Han$^3$\thanks{Keejun Han and Mun Yong Yi are corresponding authors.}
\and
Mun Yong Yi$^4$\footnotemark[1]
\\
\affiliations
$^1$Graduate School of Data Science, KAIST, Daejeon, Republic of Korea\\
$^2$Data Platform Team, SOCAR, Seoul, Republic of Korea\\
$^3$Division of Computer Engineering, Hansung University, Seoul, Republic of Korea\\
$^4$Department of Industrial and Systems Engineering, KAIST, Daejeon, Republic of Korea\\
\emails
\{handonghee, dh.kim, dybroh\}@kaist.ac.kr, lmj35021@gmail.com, keejun.han@hansung.ac.kr, munyi@kaist.ac.kr
}

\begin{document}

\maketitle

\begin{abstract}
The rise of online learning has led to the development of various knowledge tracing (KT) methods. 
However, existing methods have overlooked the problem of increasing computational cost when utilizing large graphs and long learning sequences. 
To address this issue, we introduce Dual Graph Attention-based Knowledge Tracing (DGAKT), a graph neural network model designed to leverage high-order information from subgraphs representing student-exercise-KC relationships. 
DGAKT incorporates a subgraph-based approach to enhance computational efficiency. By processing only relevant subgraphs for each target interaction, DGAKT significantly reduces memory and computational requirements compared to full global graph models. 
Extensive experimental results demonstrate that DGAKT not only outperforms existing KT models but also sets a new standard in resource efficiency, addressing a critical need that has been largely overlooked by prior KT approaches. 
\end{abstract}

\input{1.Introduction}

\input{4.Method_2}

\input{5.Experiment}
\input{7.Conclusion}

\bibliographystyle{named}
\bibliography{cite}

\end{document}

%% file: 1.Introduction.tex
\section{Introduction}
Online education has grown rapidly, prompting efforts to predict student achievement using their records. Knowledge tracing (KT) helps assess a student’s ability to complete exercises based on past interactions.\cite{kt_survey}. 
Each exercise relates to knowledge concepts (KCs), which represent the knowledge required to solve the exercise.
KT is essential for recommending suitable learning content and delivering personalized learning curriculum.

With recent advances in deep learning, deep learning based KT methods have emerged \cite{atdkt,akt,saintp,kt_survey_2023}, to track the changing knowledge state of students. Recent studies on KT have further utilized graph neural networks (GNNs) to learn the relationships between KCs and exercises-KCs by converting them into graphs \cite{self-sup-graph,gikt,pebg,gkt}.
In this context, although various graph-based KT models have been proposed, they have not adequately addressed the issue of exponentially increasing computational complexity as the size of the graph expands. Moreover, these models often overlook the potential of high-order information embedded in indirect paths by failing to effectively integrate student-exercise and exercise-KC relationships.


To overcome these limitations, it is essential to integrate student-exercise and exercise-KC relationships, enabling the utilization of high-order information embedded in indirect paths. High-order information refers to patterns and relationships that extend beyond direct connections \cite{high_order_survey}. For instance, a more detailed understanding of relationships between two nodes can emerge through indirect paths that combine multiple types of relationships. Studies in recommender systems have emphasized the importance of high-order information \cite{high_order_survey,gnn_recsys,high_order_social}. Similarly, previous studies have employed GNNs to uncover diverse connectivity patterns in graphs and to capture high-order information \cite{hosc,hae}.
In the KT field, however, research on approaches that effectively utilize such high-order information remains insufficient. Most prior studies have primarily focused on building graphs based on exercise-KC or exercise-exercise relationships, using exercise and KC node embeddings in sequence models \cite{hgkt,gkt,gikt,pebg}. These methods face significant limitations in capturing high-order information and in comprehensively integrating student-exercise and exercise-KC relationships.

\begin{figure*}[ht]
  \centering
  \includegraphics[width=13cm]{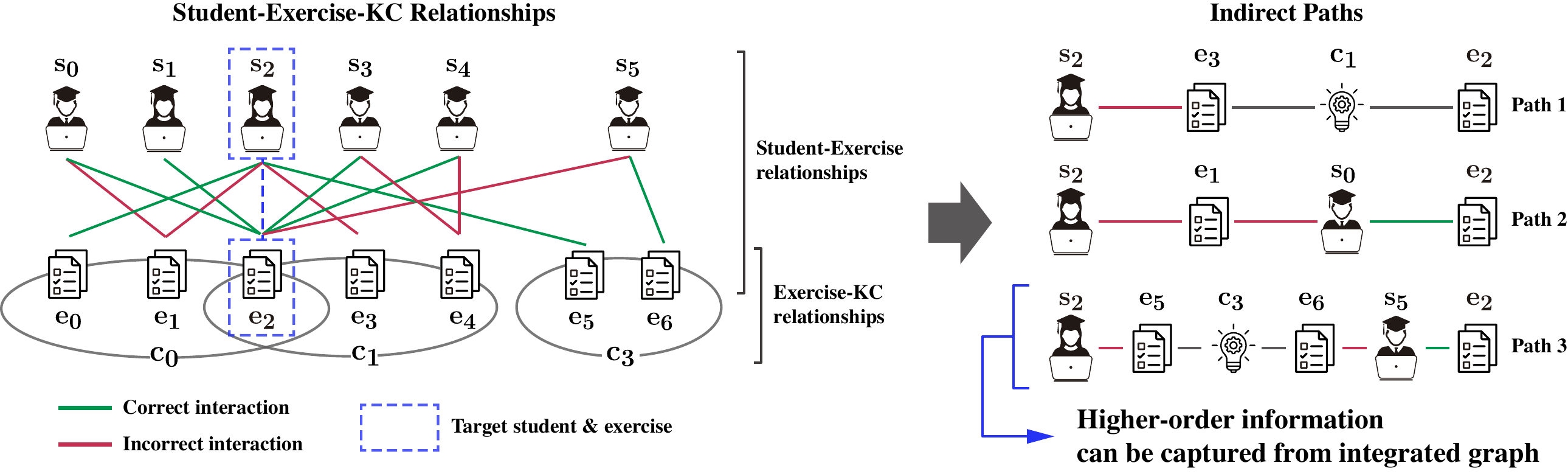}
  \caption{\label{fig:high_orer}Illustrative example of a Student-Exercise-KC graph and indirect paths between target student and exercise. Several paths that indirectly connect the target student $s_2$ to the target exercise $e_2$ can be found through multiple hops.}
\end{figure*}

As shown in Figure \ref{fig:high_orer}, higher-order information can be captured by integrating student-exercise and exercise-KC relationships, covering both direct and indirect relationships.
The three paths shown on the right side of the figure are examples of paths that may be found between the target student $s_2$ and the target exercise $e_2$.
In existing KT models, the exercise-KC relationships and other students' interactions are typically treated as separate components. Thus, cases such as \emph{path 1} and \emph{2} can be discovered, but it is difficult to find \emph{path 3}, which utilizes the student-exercise relationships of other students and exercise-KC relationships together.
\emph{path 3} shows that additional indirect paths can be discovered by combining multiple relation types. 
Consider $c_3$, which is a KC that has no overlapping KCs, the discovery of the new \emph{path 3} becomes possible only by integrating the exercise-KC and student-exercise relationships.
In this paper, we propose a method designed to combine student-exercise and exercise-KC relationships to capture high-order information from diverse paths, thereby enhancing predictive performance for student achievement.


To address the limitations of existing KT models and harness additional high-order information, we propose Dual Graph Attention-based Knowledge Tracing (DGAKT), which learns complex relationships contained in an integrated student-exercise-KC graph.
To effectively capture various graph structure patterns and handle large datasets efficiently, we utilize a subgraph-based approach.
Target subgraphs are constructed based on the interaction sequences of target students and target exercises. This approach not only prevents the model from becoming oversized but also improves its generalization ability, allowing it to infer newly emerging exercises or KCs.
DGAKT leverages graph attention networks in two aspects: 
(1) the first is the local graph attention layer, which computes the attention scores of each node's neighbors, and (2) the second is the global graph attention layer, which computes the importance of nodes in the subgraph and reflects this determination in the final subgraph representation. 
The global graph attention aggregates information from nodes that are not directly connected and can help capture high-order information. 
This dual attention layer drives the performance and interpretability of our model by deterimining the individual importance of the student, exercise, and KC nodes in the subgraph.

By focusing on a subgraph-based approach, DGAKT significantly reduces memory consumption compared to models that rely on full global graphs, as it only processes the pertinent subgraph for each target student-exercise interaction. This structure eliminates unnecessary computations and enables feasible training and deployment, even in limited-resource environments.  Such design attributes make DGAKT particularly suited for real-time inference in data-intensive educational platforms, where efficient data engineering and resource management are crucial.
Our analyses and experiments demonstrate that DGAKT not only excels in predictive performance but also substantially lowers computational overhead, proving to be a more practical and resource-efficient solution for large-scale KT tasks.


Our contributions in this work are as follows: 
\begin{itemize}
    \item We propose a novel subgraph-based knowledge tracing model that leverages higher-order information from integrated student-exercise-KC graphs.
    \item We introduce a dual graph attention structure that accounts for both local and global perspectives in the subgraph.
    \item We empirically validate the proposed approach through extensive experiments, demonstrating its superior performance compared to existing methods using multiple datasets, including large-scale datasets with over 100K students.
    \item We conduct a comprehensive analysis of DGAKT’s computational efficiency, showcasing its scalability and reduced computational requirements compared to existing KT models, making it suitable for real-world, large-scale educational applications.
\end{itemize}


%% file: 4.Method_2.tex
\section{Method \label{sec4}}

\begin{figure*}[!hbt]
  \centering
  \includegraphics[ 
  width=0.95\textwidth 
  ]{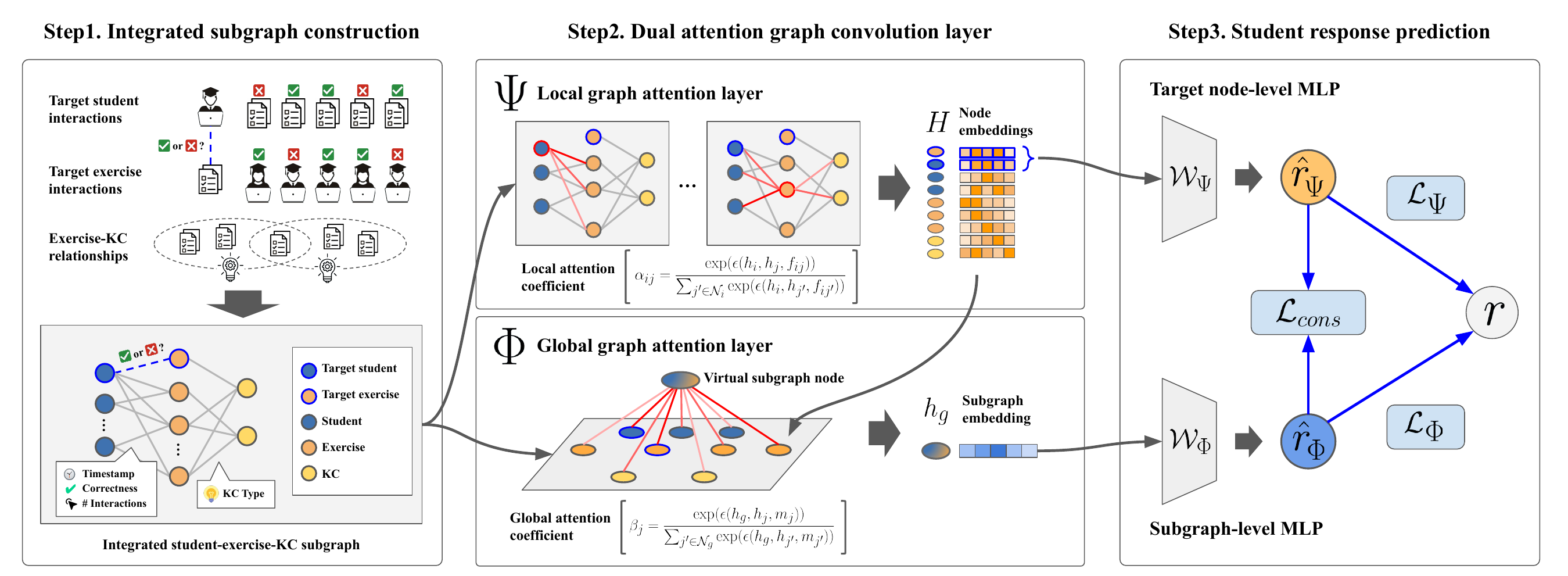}
  \caption{\label{fig:overview}The framework of the proposed model, DGAKT.}
\end{figure*}

In this section, we introduce our proposed DGAKT model and provide a detailed descriptions of its key components. 

\subsection{Overview}
Figure \ref{fig:overview} illustrates the overall framework of DGAKT, which consists of three steps for tracing student knowledge. \textbf{Step 1} constructs an integrated student-exercise-KC subgraph based on the target student's learning sequence and target exercise's interactions. \textbf{Step 2} employs dual attention graph convolution layers to construct node embeddings and a subgraph embedding. Finally, \textbf{Step 3} predicts the probability of the student correctly completing the given exercise from the node embeddings and the subgraph embedding.

Our proposed model stands out from existing models by integrating the student-exercise bipartite graph and the exercise-KC hypergraph into a single, unified graph. This integrative structure enables learning of high-order information. Additionally, our model does not rely on sequence models, such as RNN or Transformer, and instead uses only GNN layers to predict the student's response.

\subsection{Integrated subgraph construction}
To utilize higher-order information, we propose constructing a single integrated graph instead of individually learning the student-exercise graph and exercise-KC hypergraph.
Our method utilizes the interactions of multiple students and incorporates the relationships between KCs within each exercise.
Each exercise encompasses multiple KCs, which can be represented in the form of a hypergraph. Similar to previous research, we transform each hyperedge into a virtual node representing the KC to capture the characteristics of the hyperedge \cite{social_hypergraph}. 
By doing so, the properties within each KC are modeled into an embedding space instead of solely utilizing the hyperedge as a message passing path.

DGAKT utilizes subgraphs based on fixed length learning sequences. We split each student's learning sequence into non-overlapping subsequences. Student-exercise-KC subgraphs are constructed with the following three components: (1) the exercises in the target student's learning sequence, (2) the students who have interacted with the target exercise, and (3) the KCs connected to the exercises. In the subgraph, all relationships between the selected nodes are included except the target interaction. 

We extend the labeling trick proposed in previous study\cite{igmc} to label target students and exercises as 0 and 1, and neighboring students and exercises as 2 and 3, respectively. The two types of KCs related to the exercises are labeled 4 and 5, respectively. We then initialize all node embeddings as one-hot vectors with each position corresponding to a label as 1 and otherwise as 0. Through this process, each node is assigned a feature that represents the structure of the subgraph. 




\subsection{Dual attention graph convolution layer}

Our proposed model uses a unique approach during the learning process by combining the perspectives of two Graph Attention Networks (GATs) \cite{gat}. 
Our proposed model is effective for not only capturing direct connections between nodes in the integrated graph via message passing, but also incorporating signals from indirectly connected relationships through a virtual central node.

DGAKT is comprised of two key components: a local graph attention layer $\Psi$ and a global graph attention layer $\Phi$. The local graph attention layer is designed to capture the relationship between each node and its neighbors, while the global graph attention layer focuses on learning the relative importance of each node within the target subgraph. 
Since global graph attentions simultaneously aggregate representations of nodes that are not directly connected, they can capture signals that are often attenuated by traditional multi-hop message passing.


\begin{equation}
    \label{equ:dual_gat}
    \begin{aligned}
    H^l &= \Psi^l(\mathcal{G}_{sub},H^{l-1},F)\\
    h_g^l &= \Phi^l(\mathcal{G}_{sub},H^l,M)
\end{aligned} \vspace{0.5em}
\end{equation}

Equation (\ref{equ:dual_gat}) shows how the two graph attention layers generate node embeddings and a subgraph embedding. 
The $l-th$ local graph attention layer $\Psi^l$ generates a node embedding set $H^l \in \mathbb{R}^{|\mathcal{V}|\times d_h}$ from a target subgraph $\mathcal{G}_{sub}$, a previous node embedding set $H^{l-1}$, and an edge feature set $F \in \mathbb{R}^{|E|\times d_f}$. $d_h$ and $d_f$ denote the dimension of the node embedding and the dimension of the edge feature, respectively. The updated node embedding set $H^l$ is fed into the global attention layer $\Phi^l$ to create a subgraph embedding $h_g^l$. $M \in \mathbb{R}^{|\mathcal{V}|\times \mu}$ is a feature set of the edges connecting the entire node to the subgraph nodes, including the types of nodes connected. $\mu$ denotes the number of node types.

\subsubsection{Local graph attention}
The local graph attention layer computes the weights of each node's neighbors while accounting for edge features, and it also performs message passing based on the computed weights.
For local graph attention, the \emph{timestamp}, \emph{number of previous interactions} and the \emph{response} from the student-exercise interactions are combined into 3-dimensional vectors and used as the initial edge features. We limit the \emph{number of previous interactions} to a maximum of 128 and apply min-max normalization.

\begin{equation}
    \label{equ:ts_norm}
\begin{aligned}
    {ts}_\text{abs} &= |{ts}_{\tau} - ts| 
    \\[2pt]
    {ts}_\text{norm} &= 1 - \frac{{ts}_\text{abs} - min({ts}_\text{abs})}{max({ts}_\text{abs})-min({ts}_\text{abs})}
\end{aligned} \vspace{0.5em}
\end{equation}

For timestamps, as shown in Equation (\ref{equ:ts_norm}), the closer the timestamp is to the target interaction time, the closer it is to 1 and the farther it is from 0. ${ts}_{\tau}$ and ${ts}_\text{norm}$ denote the target timestamp and normalized timestamp, respectively.

EGAT \cite{egat} layers are employed to compute the attention coefficients of each node's neighbors. The attention coefficients are calculated by concatenating the edge feature vector and the two connected node vectors. Equation (\ref{equ:attn_1}) and (\ref{equ:attn_2}) calculate the attention weights in the EGAT layer:

\begin{equation}
    \label{equ:attn_1}
    \epsilon(h_{i}, h_{j}, f_{ij} ) = \text{LeakyReLU}(attn([W_{i}h_{i}||W_{j}h_{j}||W_{f}f_{ij}])
\end{equation}

\noindent where $h_{i}$ represents the features of node $i$ and $f_{ij}$ represents the features of the edge between nodes $i$ and $j$. The features of the source and destination nodes and the edge between them are transformed using weight matrices $W_i$, $W_j$, and $W_f$. An attention mechanism is then applied. $||$ denotes concatenation operation, and the function $attn$ is a linear transformer that computes the attention. Then, Equation (\ref{equ:attn_2}) is calculated as:

\begin{equation}
    \label{equ:attn_2}
     {\alpha}_{ij}
     = \cfrac {\text{exp}(\epsilon({h}_{i}, {h}_{j}, {f}_{ij} ))}{ \sum_{{j}^{\prime}\in \mathcal{N}_{i}} \text{exp}(\epsilon({h}_{i}, {h}_{{j}^{\prime}}, f_{ij^{\prime}}))}
\end{equation}

\noindent where $\mathcal{N}_i$ denotes the node group of the neighbors of node $i$. By applying the softmax function to capturing the importance of all neighboring nodes, the coefficient $\alpha_{ij}$, which denotes the attention between nodes $i$ and $j$, is obtained and used to determine the weight of information propagated between nodes $i$ and $j$ in the message passing step.

\begin{equation}
    \label{equ:multi_head1}
    h_{i}^{\prime} = \text{ELU} \left( W_{\text{local}} \concat_{k=1}^K  \sum_{j \in \mathcal{N}_i}  \alpha_{ij}^k \cdot W_{\Psi} h_j \right)
\end{equation}

Once the local attention coefficient is computed, message passing is conducted by utilizing the attention coefficient as a weight, as depicted in Equation (\ref{equ:multi_head1}). The updated node embeddings through the local graph attention layer are represented by $h_i^{\prime}$. The number of attention heads used is represented by $K$, and the attention coefficient from each attention head $k$ is represented by $\alpha_{ij}^k$. The weight for the transformation of the node embedding is represented by $W_{\Psi}$. All messages delivered from each head are concatenated into a vector and transformed by $W_{\text{local}}$. Exponential Linear Unit (ELU) \cite{elu} is adopted as the activation function.

\subsubsection{Global graph attention}
Global graph attention layer utilizes virtual nodes to represent each subgraph.
The virtual node, referred to as subgraph node $g$, is connected to all nodes in the subgraph and initialized with a zero vector. Additionally, unidirectional edges are established from nodes within subgraphs to the subgraph node. subgraph node only receives messages from other nodes to prevent oversmoothing. The features of the edges are given as one-hot vectors indicating the types of nodes connected to the edge. 
This design aims to learn the importance of each node and make the model interpretable, particularly when considering the inclusion of KC nodes, allowing for the identification of crucial KCs.

\begin{equation}
    \label{equ:attn_3}
     {\beta}_{j}
     = \cfrac {\text{exp}(\epsilon({h}_{g}, {h}_{j}, {m}_{j} ))}{ \sum_{{j}^{\prime}\in \mathcal{N}_{g}} \text{exp}(\epsilon({h}_{g}, {h}_{{j}^{\prime}}, m_{j^{\prime}}))}
\end{equation}

In Equation (\ref{equ:attn_3}), the attention coefficient between the subgraph node $g$ and node $j$ is represented by $\beta_j$. The node type of $j$ is denoted by the one-hot vector $m_j$.

\begin{equation}
    \label{equ:multi_head2}
    h_{g}^{\prime} = \text{ELU} \left( W_{\text{global}} \concat_{k=1}^K  \sum_{j \in \mathcal{N}_{g}}  \beta_{j}^k \cdot W_{\Phi} h_j \right)
\end{equation}

Equation (\ref{equ:multi_head2}) illustrates the process of updating the embedding of a subgraph node, where $h_{g}^{\prime}$ represents the updated subgraph node embedding, $\mathcal{N}_{g}$ denotes the list of nodes within the subgraph, and $W_{\Phi}$ represents the weight for the transformation of the node embedding in the global graph attention layer. All messages delivered from each attention head are concatenated into a vector and transformed by $W_{\text{global}}$.

\subsection{Student response prediction}
The two types of graph attention layers are alternately stacked $L$ times. The node embeddings and subgraph embedding from each layer are concatenated into one embedding each as shown in Equation (\ref{equ:concat}):

\begin{equation}
    \label{equ:concat}
    \begin{aligned}
    x_g &= \text{concat}(h_g^1, h_g^2, ...\:, h_g^L)\\
    x_{\tau} &= \text{concat}(h_s^1, h_s^2, ... \:, h_s^L,\:\: h_e^1, h_e^2, ... \:, h_e^L)
    \end{aligned}
\end{equation}

\noindent $h_s^l$ and $h_e^l$ denote the target student node embedding and target exercise node embeddings from the $l-th$ local graph attention layer respectively. $h_g^l$ denotes the subgraph embedding from the $l-th$ global graph attention layer. $x_g$ denotes the final subgraph embedding, $x_{\tau}$ denotes the final target node pair embeddings, and $L$ is the total number of layers. In our implementation, we used a value of two for $L$.

After obtaining the subgraph node representations, the predicted probability of correct response is derived through a fully connected layer. Equation (\ref{equ:prediction}) shows the process of obtaining the final response prediction $\hat{r}$: 

\begin{equation}
\begin{aligned}
\label{equ:prediction}
  \hat{r_\Phi} &= \sigma(\mathcal{W}_\Phi^1 (\mathcal{W}_\Phi^0 x_{g}))\\
  \hat{r_\Psi} &= \sigma(\mathcal{W}_\Psi^1 (\mathcal{W}_\Psi^0 x_{\tau}))\\
  \hat{r} &= \gamma\hat{r_\Phi}+(1-\gamma)\hat{r_\Psi}
\end{aligned}
\end{equation}

\noindent $\mathcal{W}$ denotes the parameters of the fully connected layers. A sigmoid function $\sigma$ is applied to normalize $\hat{r}$ between 0 and 1. 
We combined the two results using the hyperparameter $\gamma$.

\subsection{Model training}

To learn the local-global perspective together, we designed a loss function that considers the consistency of the two perspectives. 
The loss function is presented in Equations (\ref{equ:loss_global}), (\ref{equ:loss_local}), (\ref{equ:loss_consistency}), and (\ref{equ:loss_fin}):

\begin{equation}
\label{equ:loss_global}
    \mathcal{L}_{\Phi} = -\frac{1}{N} \sum_{i=0}^{N} {(r_{i}\log(\hat{r_\Phi}_{i}) + (1 - r_{i})\log(1 - \hat{r_\Phi}_{i}))} 
\end{equation}  

\begin{equation}
\label{equ:loss_local}
    \mathcal{L}_{\Psi} = -\frac{1}{N} \sum_{i=0}^{N} {(r_{i}\log(\hat{r_\Psi}_{i}) + (1 - r_{i})\log(1 - \hat{r_\Psi}_{i}))} 
\end{equation}  

\begin{equation}
\label{equ:loss_consistency}
    \mathcal{L}_{cons} = \sqrt{\frac{1}{N}\textstyle \sum_{i=1}^{N}(\hat{r}_{\Phi i} - \hat{r}_{\Psi i})^{2}}
\end{equation}  

\begin{equation}
\label{equ:loss_fin}
   \begin{aligned}
    \mathcal{L} = \gamma\mathcal{L}_{\Phi}+(1-\gamma)\mathcal{L}_{\Psi} + \lambda\mathcal{L}_{cons}
   \end{aligned}
\end{equation}

\noindent where $N$ represents the number of samples, $r$ represents the ground-truth response, and $\hat{r}$ denotes the predicted probability of the correct response. 
We set the training objective to minimizing the binary cross-entropy loss (BCE loss) of the response prediction. 
The global loss $\mathcal{L}_{\Phi}$ is the BCE loss between the predicted response $\hat{r_\Phi}$ from the subgraph embedding and the ground truth $r$.
The local loss $\mathcal{L}_{\Psi}$ is the BCE loss between the predicted response $\hat{r_\Psi}$ from the node embedding and the ground truth $r$.
We combined the two losses using the hyperparameter $\gamma$. 
In addition, we added a term $\mathcal{L}_{cons}$ to increase the consistency of the two perspectives, and utilize the Root Mean Square Error (RMSE) loss between $\hat{r_\Phi}$ and $\hat{r_\Psi}$. This consistency term reflects the decreasing confidence of the model when two perspectives disagree. The hyper parameter $\lambda$ is adopted to control the consistency term.
To optimize the model, an Adam optimizer \cite{adam} is used. 

%% file: 5.Experiment.tex
\section{Experiment \label{sec5}}
In this section, we present the experimental setup and results. The proposed DGAKT model is evaluated against other models to address the following research questions.


\noindent\textbf{RQ1.} Can the proposed method improve KT performance over existing baselines? 

\noindent\textbf{RQ2.} How does the local graph attention and global graph attention impact the performance of the proposed method?

\noindent\textbf{RQ3.} How do the added components and edge features affect performance?

\noindent\textbf{RQ4.} Can the proposed method generalize to new KCs and exercises

\subsection{Dataset}

The proposed method was evaluated using three datasets from distinct domains: \textbf{EdNet} \cite{ednet}, \textbf{ASSISTments2017} (ASSIST2017) \cite{assistment_dataset}, and \textbf{Junyi}\footnote{ https://pslcdatashop.web.cmu.edu/}. We included datasets comprising over 100K students, in consideration of web-based learning systems capable of serving a substantial student user base. 
The detailed statistics of datasets are presented in Table \ref{tbl:dataset}.

\begin{table}[!ht]
    \centering
    
    \resizebox{\columnwidth}{!}{\renewcommand{\arraystretch}{1}
    \begin{tabular}{l|c c c}
    \hline
    \hline
    Dataset                   & EdNet        & ASSIST2017     & Junyi \\\hline
    skill type                & Multiple      & Single         & Single     \\
    \#student                 & 117,345       & 1,709          & 162,303     \\
    \#exercise                & 13,517        & 3,162          & 722         \\
    \#skill                   & 188           & 102            & 49         \\
    \#skill per exercise      & 2.292         & 1.0            & 2.0        \\
    \#exercise type           & 7             & 16             & 257         \\
    \#interaction per student & 255.66        & 551.68         & 18.48     \\
    \#interaction             & 30,000,000    & 942,816        & 3,000,000  \\
    \% of correct responses & 66.76\% & 37.30\% & 82.80\% \\
    \hline
    \hline
    \end{tabular}
    }
    \caption{Statistics of datasets.\label{tbl:dataset}}

\end{table}

\subsection{Baselines}
To demonstrate the effectiveness of our DGAKT model, we compared it with the most widely used baselines, and recent methods related to our work from three different perspectives: (1) RNN-based models, such as DKT\cite{dkt}, AT-DKT\cite{atdkt} and DKVMN\cite{dkvmn}, (2) an attention-based models, such as AKT\cite{akt}, SAKT\cite{sakt} and SAINT\cite{saint}, and (3) graph-based models, such as GKT\cite{gkt}, IGMC\cite{igmc}, PEBG\cite{pebg} and DGEKT\cite{dgekt}. To compare the effect of utilizing an integrated graph, we measured the performance of an IGMC variant that includes KC nodes in the subgraphs (IGMC-KC).

\subsection{Experimental Settings}

\subsubsection{Evaluation Metric} 

To compare the prediction performance of DGAKT with other baseline methods, the \emph{Accuracy} (ACC) and the \emph{Area Under the ROC Curve} (AUC) were used as evaluation metrics. These metrics are widely used in KT studies \cite{kt_survey,pykt}.

\subsubsection{Train Setting} 

All datasets were split chronologically into training (60\%), validation (20\%), and testing (20\%) sets. We used DGL \cite{dgl} and PyTorch \cite{pytorch} for implementation.
More detailed experimental settings and datasets can be found in our public repository.

\begin{table}[t!]
    \centering
    
    \resizebox{\columnwidth}{!}{
    \renewcommand{\arraystretch}{1.2}
    \begin{tabular}{c|c c c c c c}
    \hline\hline
    Dataset          & \multicolumn{2}{c}{EdNet}                         & \multicolumn{2}{c}{ASSIST2017}                 & \multicolumn{2}{c}{Junyi}          \\ \cline{2-3} \cline{4-5} \cline{6-7}\hline
    Metric        & \multicolumn{1}{c}{ACC} & \multicolumn{1}{c}{AUC} & \multicolumn{1}{c}{ACC} & \multicolumn{1}{c}{AUC} & \multicolumn{1}{c}{ACC} & \multicolumn{1}{c}{AUC} \\\hline\hline
    DKT           & \ccell{0.6868}{0.0006}              & \ccell{0.6771}{0.0012}              & \ccell{0.6789}{0.0003}             & \ccell{0.7005}{0.0008}             & \ccell{0.8459}{0.0004}	         & \ccell{0.7465}{0.0009}                   \\
    AT-DKT         & 0.6909	          & 0.6928	         & 0.6759	        & 0.7022         & 0.8465	        & 0.7449  \\
    DKVMN         & \ccell{0.6859}{0.0005}              & \ccell{0.6732}{0.0002}              & \ccell{0.6795}{0.0006}             & \ccell{0.6952}{0.0012}             & \ccell{0.8449}{0.0011}	         & \ccell{0.7424}{0.0013}                   \\  \hline
    AKT           & \ccell{0.6938}{0.0021}	            & \ccell{0.7145}{0.0036}	          & \ccell{0.6891}{0.0014}             & \ccell{0.7263}{0.0009}             & \ccell{\underline{0.8503}}{0.0019} & \ccell{0.7877}{0.0042}                    \\
    SAKT          & \ccell{0.6355}{0.0130}              & \ccell{0.6407}{0.0091}              & \ccell{0.6575}{0.0040}             & \ccell{0.6742}{0.0033}             & \ccell{0.8448}{0.0028}             & \ccell{0.7439}{0.0035}                    \\
    SAINT         & \ccell{\underline{0.7381}}{0.0033}  & \ccell{0.7724}{0.0043}              & \ccell{0.6718}{0.0024}             & \ccell{0.6951}{0.0032}             & \ccell{0.8392}{0.0018}             & \ccell{0.7119}{0.0023}                    \\ \hline
    GKT           & \ccell{0.7118}{0.0031}              & \ccell{0.6697}{0.0021}              & \ccell{0.6858}{0.0016}             & \ccell{0.7125}{0.0032}             & \ccell{0.8459}{0.0020}             & \ccell{0.7470}{0.0039}              \\
    PEBG          & \ccell{0.6169}{0.0015}              & \ccell{0.6438}{0.0026}              & \ccell{0.7061}{0.0033}             & \ccell{0.7644}{0.0031}             & \ccell{0.8553}{0.0022}             & \ccell{0.7972}{0.0019}                    \\
    DGEKT         & \ccell{0.6956}{0.0002}              & \ccell{0.7392}{0.0009}              & \ccell{0.6799}{0.0043}             & \ccell{0.7346}{0.0027}             & \ccell{0.8433}{0.0023}             & \ccell{0.8304}{0.0015}                    \\ \hline
    IGMC          & \ccell{0.6209}{0.0023}              & \ccell{0.6184}{0.0014}              & \ccell{0.6898}{0.0097}             & \ccell{0.7572}{0.0021}             & \ccell{0.8443}{0.0016}             & \ccell{\underline{0.8796}}{0.0009}  \\
    IGMC-KC       & \ccell{0.7348}{0.0022}              & \ccell{\underline{0.7864}}{0.0015}  & \ccell{\underline{0.7268}}{0.0015} &\ccell{\underline{0.7985}}{0.0018}  & \ccell{0.8454}{0.0013}             & \ccell{0.8774}{0.0023}              \\\hline
    \textbf{DGAKT(Ours)}   & \ccell{\textbf{0.7630}}{0.0019}     & \ccell{\textbf{0.8188}}{0.0028}     & \ccell{\textbf{0.8151}}{0.0039}    & \ccell{\textbf{0.8994}}{0.0045}    & \ccell{\textbf{0.8576}}{0.0011}    & \ccell{\textbf{0.9122}}{0.0017}     \\ \hline
    Improvement   & 3.26\%               & 3.95\%               & 10.83\%              & 11.22\%            & 0.85\%            & 3.57\%             \\
    \hline\hline 
    \end{tabular}
    }
    \caption{Accuracy and AUC performance of baselines and DGAKT.\label{tbl:result}}
    \end{table}

\subsection{Experiment Results}

\subsubsection{Overall Performance}

We conducted experiments to evaluate our model against baselines to answer \textbf{RQ1}. Table \ref{tbl:result} shows the overall performance outcomes. The best results are highlighted in bold, and the second best results are underlined. Each experiment was repeated five times and the \emph{mean} is reported. T-tests confirmed that the performance differences between DGAKT and best baselines were all significant at a p-value $<$ 0.01.

These results show that the proposed model achieved superior performance across the datasets in ACC and AUC, outperforming all of the baselines. 
Additionally, the IGMC-KC, which incorporates student-exercise interactions and exercise-KC relationships, performed the best among all of the baseline models in most cases, further supporting the effectiveness of utilizing high-order information for KT, as we proposed. 
Furthermore, our proposed model shows a greater improvement on the ASSIST2017 dataset than others.
ASSIT2017 is the dataset with the highest average number of student interactions, as presented in Table \ref{tbl:dataset}, which indicates a denser graph. 
In such a dense graph, more indirect paths can be found between node pairs, which suggests that rich high-order information can be utilized. 
Due to these characteristics, we can interpret that our proposed method works effectively. 
We can also see that the performance improvements on other datasets are correlated with the average number of interactions per student. These results support that our model can effectively utilize high-order information.


\begin{table}[!t]
\centering
\fontsize{8}{8}\selectfont
\resizebox{0.9\columnwidth}{!}{
\renewcommand{\arraystretch}{1.2}
\begin{tabular}{c|c|c|c}  
    \hline
    Variant & Local attention & Global attention & Knowledge concept  \\ \hline
    V1 & \xmark & \xmark & \cmark \\ 
    V2 & \cmark & \xmark & \cmark \\ 
    V3  & \cmark & \cmark & \xmark \\ \hline
    DGAKT  & \cmark & \cmark & \cmark \\
    \hline
\end{tabular}
}
\caption{\label{tbl:ablation_1}Variants of DGAKT by components. }
\vspace{-0.3cm}
\end{table}

\begin{figure}[!t]
  \centering
  \includegraphics[width=0.9\columnwidth]{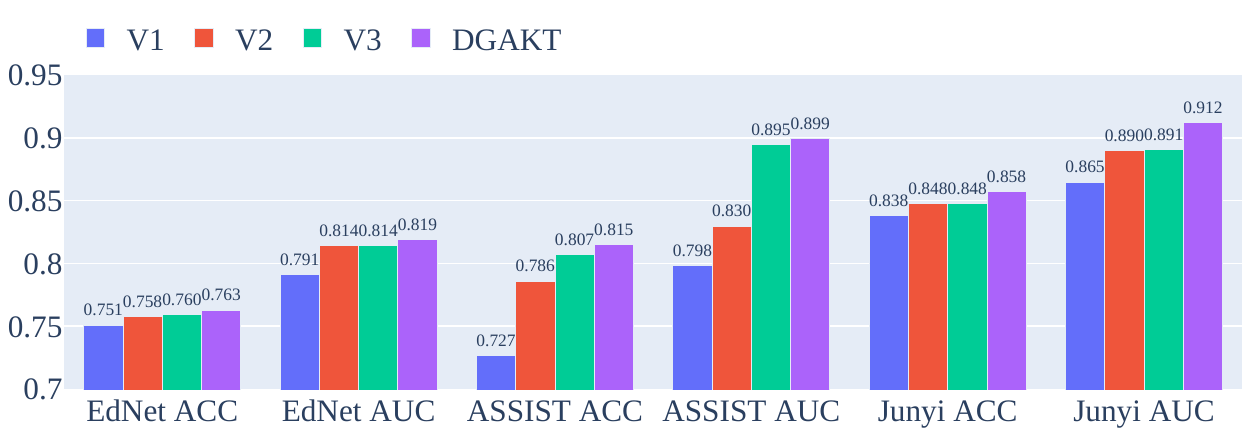}
  \caption{\label{fig:ablation_1}Ablation study by DGAKT components.}
\end{figure}

\begin{table}[!t]
\centering
\resizebox{0.9\columnwidth}{!}{
\fontsize{8}{8}\selectfont
\renewcommand{\arraystretch}{1.2}
\begin{tabular}{c|c|c|c}  
    \hline
    Variant & Timestamp & \#previous interactions & Response  \\ \hline
    V4 & \xmark & \cmark & \cmark \\ 
    V5 & \cmark & \xmark & \cmark \\ 
    V6  & \xmark & \xmark & \cmark \\ \hline
    DGAKT  & \cmark & \cmark & \cmark \\
    \hline
\end{tabular}
}
\caption{\label{tbl:ablation_2}Variants of DGAKT by edge features. }
\vspace{-0.3cm}
\end{table}

\begin{figure}[!t]
  \centering
  \includegraphics[width=0.9\columnwidth]{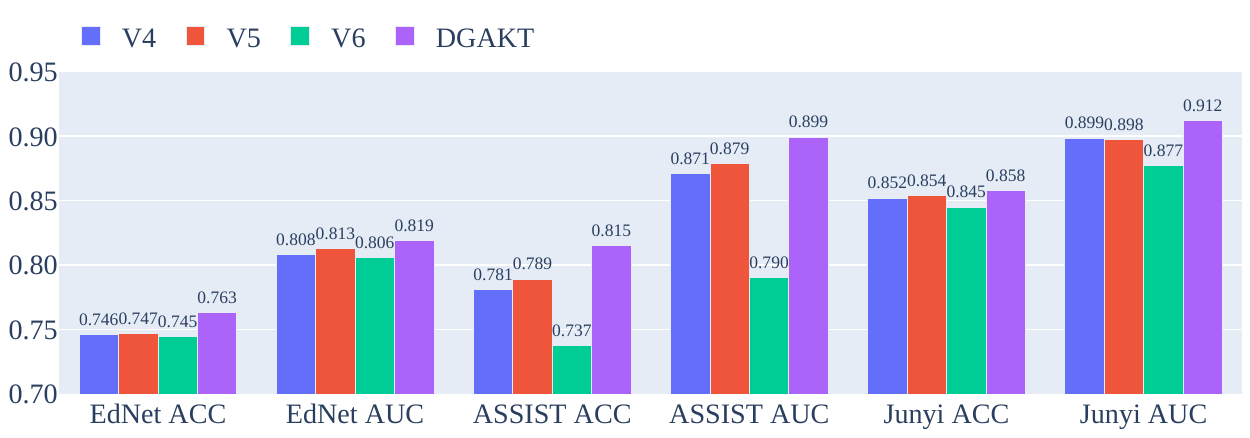}
  \caption{\label{fig:ablation_2}Ablation study by edge features.}
\vspace{-0.3cm}
\end{figure}

\subsubsection{Ablation study}
To answer \textbf{RQ2} and \textbf{RQ3}, ablation studies were conducted to evaluate DGAKT by testing three variants excluding key components (V1-V3) and three variants removing specific edge features (V4-V6), as shown in Table \ref{tbl:ablation_1} and \ref{tbl:ablation_2}. 
The results of the ablation study, as presented in Figure \ref{fig:ablation_1} and Figure \ref{fig:ablation_2}, indicate that all of the proposed variants showed inferior performance compared to the original model.

\subsubsection{Unseen cases}

We further assess the ability of the proposed model to achieve robust predictive performance in unseen cases (\textbf{RQ4}). 
In online education services, new courses, topics, and problems are often introduced. To verify that our model can infer robustly to these changes, we experimented with inferring on types of problems not seen in training.
Prior research has demonstrated that the IGMC, which utilizes inductive properties, exhibits advantages in unseen case inference \cite{igmc}. 
To determine if our proposed model also possesses these advantages, we compared the performance of DGAKT and IGMC-KC in unseen cases.
We divided the two datasets, EdNet and ASSIST2017, into train and test sets by problem type. For example, the EdNet dataset was partitioned into a train and test sets, with the train set comprising of exercise types \{1, 2, 3, 4\} and a test set comprising of exercise types \{5, 6, 7\}.  For each dataset, we created three train-test set pairs, and the problem types in the test set were not exposed during the training phase. 


\begin{table}[!ht]
    \centering
    \renewcommand{\arraystretch}{1}
    \resizebox{\columnwidth}{!}{
    \begin{tabular}{c|c c c c c c}
        \hline\hline
        Dataset          & \multicolumn{2}{c}{ASSIST2017\_A}  & \multicolumn{2}{c}{ASSIST2017\_B}  & \multicolumn{2}{c}{ASSIST2017\_C}   \\ \cline{2-3} \cline{4-5} \cline{6-7}\hline
        Metric         & \multicolumn{1}{c}{ACC} & \multicolumn{1}{c}{AUC} & \multicolumn{1}{c}{ACC} & \multicolumn{1}{c}{AUC} & \multicolumn{1}{c}{ACC} & \multicolumn{1}{c}{AUC} \\ \hline\
        IGMC-KC        & 0.8132	          & 0.8442	         & 0.7758	        & 0.8445         & 0.7954	        & 0.8363  \\
        \textbf{DGAKT}          & \textbf{0.8708}  & \textbf{0.9480}  & \textbf{0.8610}  &\textbf{0.9444} & \textbf{0.8557}  &\textbf{0.9394}  \\\hline\hline
\end{tabular}
}
\caption{\label{tbl:unseen_assit} Unseen case performance in ASSIST2017.}
\vspace{-0.5cm}
\end{table}

\begin{table}[!ht]
    \centering
    \renewcommand{\arraystretch}{1.}
    \resizebox{\columnwidth}{!}{%
    \begin{tabular}{c|c c c c c c}
        \hline\hline 
        Dataset          & \multicolumn{2}{c}{EdNet\_A}  & \multicolumn{2}{c}{EdNet\_B}  & \multicolumn{2}{c}{EdNet\_C}   \\ \cline{2-3} \cline{4-5} \cline{6-7}\hline
        Metric         & \multicolumn{1}{c}{ACC} & \multicolumn{1}{c}{AUC} & \multicolumn{1}{c}{ACC} & \multicolumn{1}{c}{AUC} & \multicolumn{1}{c}{ACC} & \multicolumn{1}{c}{AUC} \\\hline
        IGMC-KC        & 0.7657	          & 0.7949	         & 0.7149	        & 0.7608         & 0.7576	        & 0.7755  \\
        \textbf{DGAKT}          & \textbf{0.7716}  & \textbf{0.8015}  & \textbf{0.7225}  &\textbf{0.7757} & \textbf{0.7631}  &\textbf{0.7886}  \\ \hline\hline
\end{tabular}
}
\caption{\label{tbl:unseen_ednet} Unseen case performance in EdNet.}
\end{table}

Tables \ref{tbl:unseen_assit} and \ref{tbl:unseen_ednet} present the results for ASSIST2017 and EdNet, respectively.
The proposed model outperformed IGMC-KC on all of the datasets, confirming that it is more robust to unseen cases. Our model maintains high performance even when faced with new KCs, which are not simply new exercises, suggesting that it would be effective in real-world scenarios where new KCs are encountered.

\section{Analysis \label{sec6}}
In this section, we provide a deeper analysis of the properties of DGAKT. We approached this analysis from two perspectives: time and space complexity of the model and interpretability of the model. These perspectives are crucial for determining the effectiveness of our model and its applicability to real-world online education services.




\subsection{Time \& Space Complexity}
We analyzed the time and space complexity to assess whether our model can work efficiently with large amounts of data.
KT requires efficiency in terms of time and space complexity; however, it has rarely been studied explicitly \cite{dgmn,enkt}. 
Table \ref{tbl:complexity} shows the time complexities and space complexities of existing KT models and of the proposed model.

\begin{table}[!ht]
    \centering
    \resizebox{0.8\columnwidth}{!}{
    \renewcommand{\arraystretch}{1.3}
    \begin{tabular}{c|c c}  
        \hline
        Model   & Time complexity & Space complexity  \\ \hline
        DKT     & $\mathcal{O}(n\cdot d)$  &  $\mathcal{O}(n\cdot d)$  \\ 
        DKVMN   & $\mathcal{O}(n\cdot d)$  &  $\mathcal{O}(|C|\cdot d+ n\cdot d)$  \\ 
        AKT     & $\mathcal{O}(n^2\cdot d)$  &  $\mathcal{O}(n \cdot d  +d^2 )$  \\ 
        SAKT    & $\mathcal{O}(n^2\cdot d)$  &  $\mathcal{O}(n \cdot d  +d^2 )$  \\ 
        SAINT   & $\mathcal{O}(n^2\cdot d)$  &  $\mathcal{O}(n \cdot d +d^2)$  \\ 
        GKT     & $\mathcal{O}(|Q|^2\cdot d)$  &  $\mathcal{O}(|C|^2 \cdot d + n\cdot d)$  \\ 
        DGEKT   & $\mathcal{O}(|Q|^2 + |Q|\cdot d )$  &  $\mathcal{O}(|Q|^2 \cdot d + n\cdot d)$  \\ \hline
        \textbf{DGAKT}   & $\mathcal{O}(n\cdot d^2 + n^2\cdot d)$  &  $\mathcal{O}(n^2 + n \cdot d + d^2)$  \\
        \hline
    \end{tabular}
    }
    \caption{Time and space complexities of KT models.\label{tbl:complexity}}
    
    \end{table}

Our approach improves efficiency in both time and space complexity compared to GKT and DGEKT, which suffer from quadratic complexity issues with the number of exercises. Unlike GKT, which has high training time due to its \( |Q|^2 \) time complexity, and DGEKT, which requires significant GPU memory due to its adjacency matrix, our method processes data in subgraph units, making it more scalable. Additionally, our model benefits from parallelized message passing, reducing computational costs and enhancing applicability in online tutoring systems.





\begin{figure}[ht]
  \centering
  \includegraphics[width=.9\columnwidth]{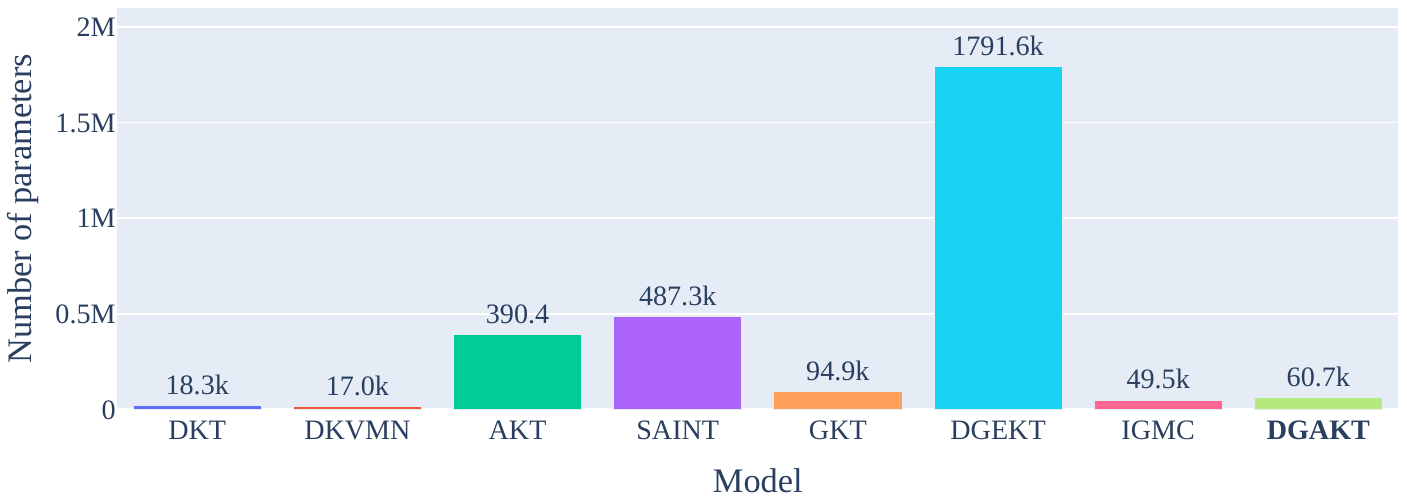}
  \caption{\label{fig:num_parameter}Comparing the number of parameters of KT models.}
\end{figure}

We compared model parameters to assess both performance and computational cost. As shown in Figure \ref{fig:num_parameter}, DGEKT requires the most parameters due to its transductive GNN-based design, while SAINT has a high parameter count from its transformer architecture. In contrast, our proposed DGAKT model has 60k parameters—higher than traditional KT models like DKT or DKVMN but reasonable given its superior performance.



\begin{figure}[!ht]
  \centering
  \includegraphics[width=0.9\columnwidth]{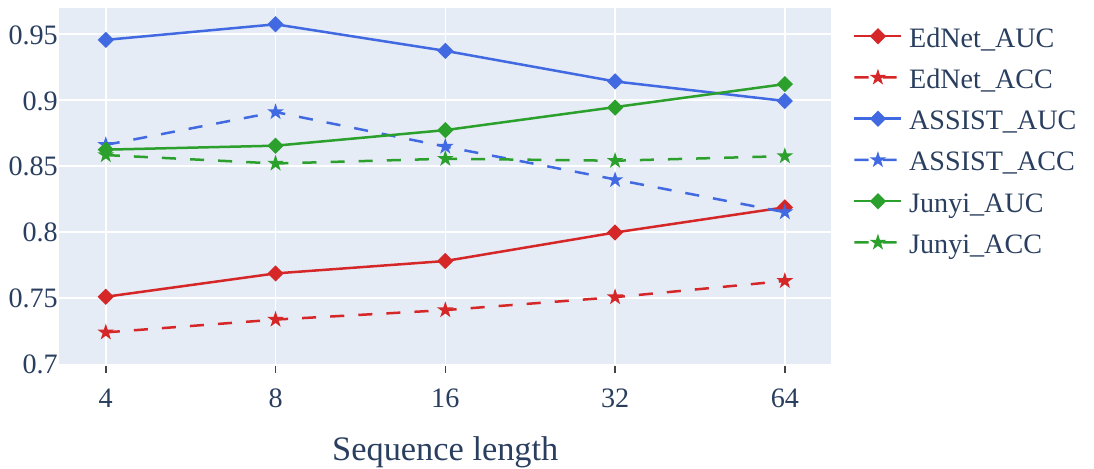}
  \caption{\label{subsequence_len}Performance changes with subsequence length}
\end{figure}

Previous research has rarely explored the impact of subsequence length on performance, though it significantly affects computational cost. Our results show that ASSIST2017 performed best at length 8, while Junyi and EdNet improved with longer sequences. Notably, our model maintained strong performance even with short subsequences (4 or 8) as shown in Figure \ref{subsequence_len}, highlighting its efficiency and applicability for students with fewer interactions in online tutoring systems.

\subsection{Interpretability}
\begin{figure}[!ht]
  \centering
  \includegraphics[width=0.95\columnwidth]{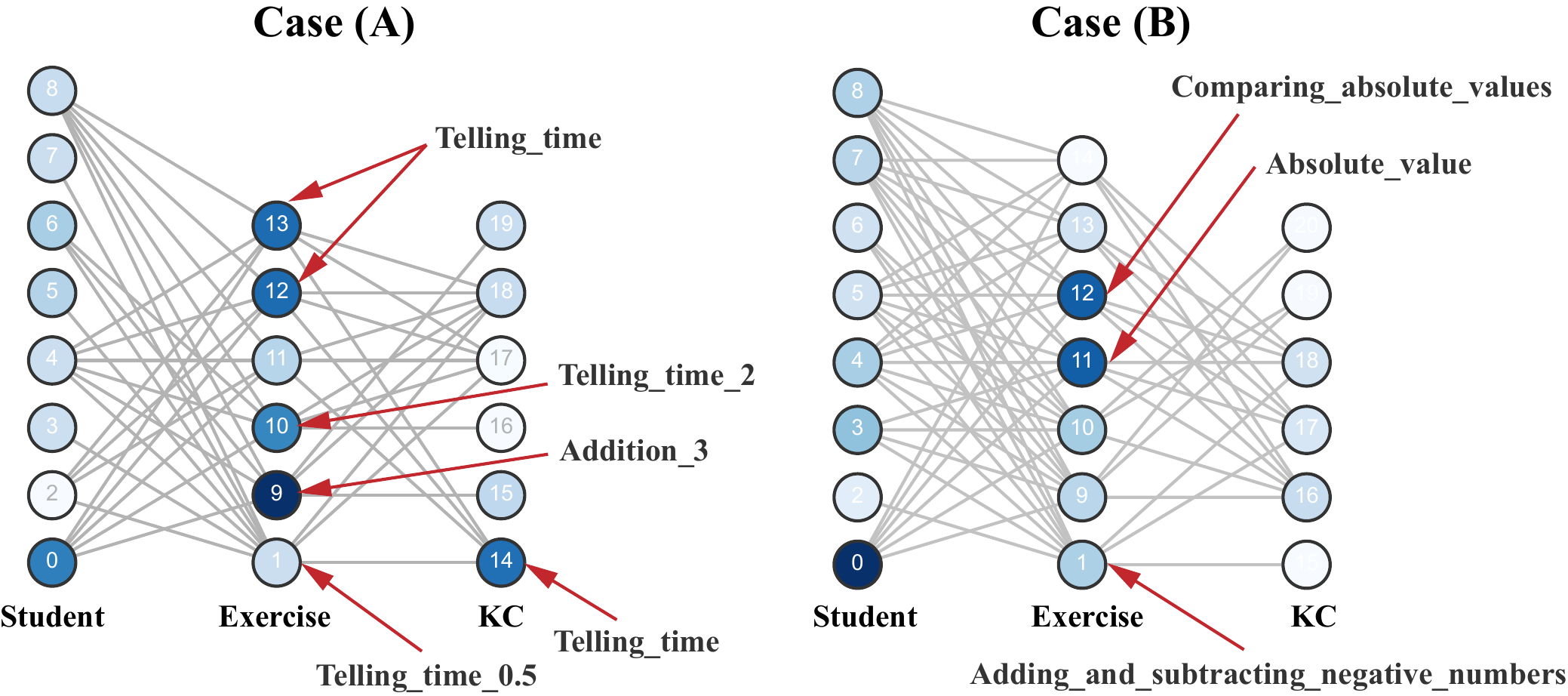}
  \caption{\label{fig:case_study}Subgraph visualization with global attention score.}
\end{figure}

DGAKT uses global attention to learn node importance within subgraphs, enabling intuitive model interpretation. To validate this, we conducted a case study using the Junyi dataset (Figure \ref{fig:case_study}), where darker blue nodes indicate higher attention scores. Key exercises and KCs are labeled, with a subsequence length of eight for clarity.

In both cases, we observe that the exercises that were more relevant to the target exercise received higher attention scores. In Case (A), the attention score of the KC related with the target exercise was also high. 
From the original data, we found that there were pre-requisite relationships between the target exercise and exercises with the higher attention scores. In Case (B), we found that \emph{Adding\_and\_subtracting\_negative\_numbers} is a pre-requisite of \emph{Absolute\_value} while \emph{Absolute\_value} is a pre-requisite of \emph{Comparing\_absolute\_values}, which is notable because we did not explicitly use pre-requisite relationships during training.
Our global graph attention-based method can help tutors to understand these important relationships during online learning. 


%% file: 7.Conclusion.tex
\section{conclusion \label{sec7}}
In this study, we proposed DGAKT, a graph-based KT model that integrates an exercise-KC hypergraph and a student-exercise graph to capture high-order information efficiently. By leveraging dual graph attention on subgraphs, DGAKT reduces computational costs while dynamically adapting to new exercises and KCs. It outperforms existing models across multiple datasets and enhances interpretability by identifying key nodes in learning patterns.


Future research could integrate LLMs like Llama \cite{llama} with graph-based KT models to enhance interpretability and accuracy while maintaining efficiency. Adapting techniques from recommendation systems \cite{allmrec,data_llm_rec}, such as fine-tuning and collaborative filtering, may further optimize LLMs for KT. Exploring LLM integration with minimal computational cost could enable scalable, real-time knowledge tracing.


\newpage

%% file: 0.main.bbl
\begin{thebibliography}{}

\bibitem[\protect\citeauthoryear{Abdelrahman and Wang}{2022}]{dgmn}
Ghodai Abdelrahman and Qing Wang.
\newblock Deep graph memory networks for forgetting-robust knowledge tracing.
\newblock {\em IEEE Transactions on Knowledge and Data Engineering}, pages 1--13, 2022.

\bibitem[\protect\citeauthoryear{Abdelrahman \bgroup \em et al.\egroup }{2023a}]{kt_survey}
Ghodai Abdelrahman, Qing Wang, and Bernardo Nunes.
\newblock Knowledge tracing: A survey.
\newblock {\em ACM Comput. Surv.}, 55(11), feb 2023.

\bibitem[\protect\citeauthoryear{Abdelrahman \bgroup \em et al.\egroup }{2023b}]{kt_survey_2023}
Ghodai Abdelrahman, Qing Wang, and Bernardo Nunes.
\newblock Knowledge tracing: A survey.
\newblock {\em ACM Comput. Surv.}, 55(11), feb 2023.

\bibitem[\protect\citeauthoryear{Barron}{2017}]{elu}
Jonathan~T. Barron.
\newblock Continuously differentiable exponential linear units.
\newblock {\em CoRR}, abs/1704.07483, 2017.

\bibitem[\protect\citeauthoryear{Choi \bgroup \em et al.\egroup }{2020a}]{saint}
Youngduck Choi, Youngnam Lee, Junghyun Cho, Jineon Baek, Byungsoo Kim, Yeongmin Cha, Dongmin Shin, Chan Bae, and Jaewe Heo.
\newblock Towards an appropriate query, key, and value computation for knowledge tracing.
\newblock In {\em Proceedings of the Seventh ACM Conference on Learning @ Scale}, L@S '20, page 341–344, New York, NY, USA, 2020. Association for Computing Machinery.

\bibitem[\protect\citeauthoryear{Choi \bgroup \em et al.\egroup }{2020b}]{ednet}
Youngduck Choi, Youngnam Lee, Dongmin Shin, Junghyun Cho, Seoyon Park, Seewoo Lee, Jineon Baek, Chan Bae, Byungsoo Kim, and Jaewe Heo.
\newblock Ednet: A large-scale hierarchical dataset in education.
\newblock In Ig~Ibert Bittencourt, Mutlu Cukurova, Kasia Muldner, Rose Luckin, and Eva Mill{\'a}n, editors, {\em \rc{Artificial Intelligence in Education}}, pages 69--73, Cham, 2020. Springer International Publishing.

\bibitem[\protect\citeauthoryear{Cui \bgroup \em et al.\egroup }{2024}]{dgekt}
Chaoran Cui, Yumo Yao, Chunyun Zhang, Hebo Ma, Yuling Ma, Zhaochun Ren, Chen Zhang, and James Ko.
\newblock Dgekt: A dual graph ensemble learning method for knowledge tracing.
\newblock {\em ACM Trans. Inf. Syst.}, 42(3), jan 2024.

\bibitem[\protect\citeauthoryear{Cully and Demiris}{2020}]{assistment_dataset}
Antoine Cully and Yiannis Demiris.
\newblock Online knowledge level tracking with data-driven student models and collaborative filtering.
\newblock {\em IEEE Transactions on Knowledge and Data Engineering}, 32(10):2000--2013, 2020.

\bibitem[\protect\citeauthoryear{Duval and Malliaros}{2022}]{hosc}
Alexandre Duval and Fragkiskos Malliaros.
\newblock Higher-order clustering and pooling for graph neural networks.
\newblock In {\em Proceedings of the 31st ACM International Conference on Information \& Knowledge Management}, CIKM '22, page 426–435, New York, NY, USA, 2022. Association for Computing Machinery.

\bibitem[\protect\citeauthoryear{Gao \bgroup \em et al.\egroup }{2022}]{gnn_recsys}
Chen Gao, Xiang Wang, Xiangnan He, and Yong Li.
\newblock Graph neural networks for recommender system.
\newblock In {\em Proceedings of the Fifteenth ACM International Conference on Web Search and Data Mining}, WSDM '22, page 1623–1625, New York, NY, USA, 2022. Association for Computing Machinery.

\bibitem[\protect\citeauthoryear{Gao \bgroup \em et al.\egroup }{2023}]{high_order_survey}
Chen Gao, Yu~Zheng, Nian Li, Yinfeng Li, Yingrong Qin, Jinghua Piao, Yuhan Quan, Jianxin Chang, Depeng Jin, Xiangnan He, and Yong Li.
\newblock A survey of graph neural networks for recommender systems: Challenges, methods, and directions.
\newblock {\em ACM Trans. Recomm. Syst.}, 1(1), mar 2023.

\bibitem[\protect\citeauthoryear{Ghosh \bgroup \em et al.\egroup }{2020}]{akt}
Aritra Ghosh, Neil Heffernan, and Andrew~S. Lan.
\newblock Context-aware attentive knowledge tracing.
\newblock In {\em Proceedings of the 26th ACM SIGKDD International Conference on Knowledge Discovery \&; Data Mining}, KDD '20, page 2330–2339, New York, NY, USA, 2020. Association for Computing Machinery.

\bibitem[\protect\citeauthoryear{Huang \bgroup \em et al.\egroup }{2021}]{high_order_social}
Chao Huang, Huance Xu, Yong Xu, Peng Dai, Lianghao Xia, Mengyin Lu, Liefeng Bo, Hao Xing, Xiaoping Lai, and Yanfang Ye.
\newblock Knowledge-aware coupled graph neural network for social recommendation.
\newblock {\em Proceedings of the AAAI Conference on Artificial Intelligence}, 35(5):4115--4122, May 2021.

\bibitem[\protect\citeauthoryear{Kim \bgroup \em et al.\egroup }{2024}]{allmrec}
Sein Kim, Hongseok Kang, Seungyoon Choi, Donghyun Kim, Minchul Yang, and Chanyoung Park.
\newblock Large language models meet collaborative filtering: An efficient all-round llm-based recommender system.
\newblock {\em arXiv preprint arXiv:2404.11343}, 2024.

\bibitem[\protect\citeauthoryear{Kingma and Ba}{2015}]{adam}
Diederik~P. Kingma and Jimmy Ba.
\newblock Adam: A method for stochastic optimization.
\newblock In Yoshua Bengio and Yann LeCun, editors, {\em \rc{International Conference on Learning Representations}}, 2015.

\bibitem[\protect\citeauthoryear{Li \bgroup \em et al.\egroup }{2023}]{hae}
Jianxin Li, Hao Peng, Yuwei Cao, Yingtong Dou, Hekai Zhang, Philip~S. Yu, and Lifang He.
\newblock Higher-order attribute-enhancing heterogeneous graph neural networks.
\newblock {\em IEEE Transactions on Knowledge and Data Engineering}, 35(1):560--574, 2023.

\bibitem[\protect\citeauthoryear{Lin \bgroup \em et al.\egroup }{2024}]{data_llm_rec}
Xinyu Lin, Wenjie Wang, Yongqi Li, Shuo Yang, Fuli Feng, Yinwei Wei, and Tat-Seng Chua.
\newblock Data-efficient fine-tuning for llm-based recommendation.
\newblock In {\em Proceedings of the 47th International ACM SIGIR Conference on Research and Development in Information Retrieval}, pages 365--374, 2024.

\bibitem[\protect\citeauthoryear{Liu \bgroup \em et al.\egroup }{2020}]{pebg}
Yunfei Liu, Yang Yang, Xianyu Chen, Jian Shen, Haifeng Zhang, and Yong Yu.
\newblock Improving knowledge tracing via pre-training question embeddings.
\newblock In {\em \rc{Proceedings of the Twenty-Ninth International Joint Conference on Artificial Intelligence}}, \rc{IJCAI'20}, 2020.

\bibitem[\protect\citeauthoryear{Liu \bgroup \em et al.\egroup }{2022}]{pykt}
Zitao Liu, Qiongqiong Liu, Jiahao Chen, Shuyan Huang, Jiliang Tang, and Weiqi Luo.
\newblock pykt: A python library to benchmark deep learning based knowledge tracing models.
\newblock In S.~Koyejo, S.~Mohamed, A.~Agarwal, D.~Belgrave, K.~Cho, and A.~Oh, editors, {\em Advances in Neural Information Processing Systems}, volume~35, pages 18542--18555. Curran Associates, Inc., 2022.

\bibitem[\protect\citeauthoryear{Liu \bgroup \em et al.\egroup }{2023}]{atdkt}
Zitao Liu, Qiongqiong Liu, Jiahao Chen, Shuyan Huang, Boyu Gao, Weiqi Luo, and Jian Weng.
\newblock Enhancing deep knowledge tracing with auxiliary tasks.
\newblock In {\em Proceedings of the ACM Web Conference 2023}, WWW '23, page 4178–4187, New York, NY, USA, 2023. Association for Computing Machinery.

\bibitem[\protect\citeauthoryear{Nakagawa \bgroup \em et al.\egroup }{2019}]{gkt}
Hiromi Nakagawa, Yusuke Iwasawa, and Yutaka Matsuo.
\newblock Graph-based knowledge tracing: Modeling student proficiency using graph neural network.
\newblock In {\em 2019 IEEE/WIC/ACM International Conference on Web Intelligence (WI)}, pages 156--163, 2019.

\bibitem[\protect\citeauthoryear{Pandey and Karypis}{2019}]{sakt}
Shalini Pandey and George Karypis.
\newblock A self-attentive model for knowledge tracing.
\newblock In {\em 12th International Conference on Educational Data Mining, EDM 2019}, pages 384--389. International Educational Data Mining Society, 2019.

\bibitem[\protect\citeauthoryear{Paszke \bgroup \em et al.\egroup }{2019}]{pytorch}
Adam Paszke, Sam Gross, Francisco Massa, Adam Lerer, James Bradbury, Gregory Chanan, Trevor Killeen, Zeming Lin, Natalia Gimelshein, Luca Antiga, Alban Desmaison, Andreas Kopf, Edward Yang, Zachary DeVito, Martin Raison, Alykhan Tejani, Sasank Chilamkurthy, Benoit Steiner, Lu~Fang, Junjie Bai, and Soumith Chintala.
\newblock Pytorch: An imperative style, high-performance deep learning library.
\newblock In H.~Wallach, H.~Larochelle, A.~Beygelzimer, F.~d\textquotesingle Alch\'{e}-Buc, E.~Fox, and R.~Garnett, editors, {\em Advances in Neural Information Processing Systems 32}, pages 8024--8035. Curran Associates, Inc., 2019.

\bibitem[\protect\citeauthoryear{Piech \bgroup \em et al.\egroup }{2015}]{dkt}
Chris Piech, Jonathan Bassen, Jonathan Huang, Surya Ganguli, Mehran Sahami, Leonidas~J Guibas, and Jascha Sohl-Dickstein.
\newblock Deep knowledge tracing.
\newblock {\em Advances in neural information processing systems}, 28, 2015.

\bibitem[\protect\citeauthoryear{Shin \bgroup \em et al.\egroup }{2021}]{saintp}
Dongmin Shin, Yugeun Shim, Hangyeol Yu, Seewoo Lee, Byungsoo Kim, and Youngduck Choi.
\newblock Saint+: Integrating temporal features for ednet correctness prediction.
\newblock In {\em LAK21: 11th International Learning Analytics and Knowledge Conference}, LAK21, page 490–496, New York, NY, USA, 2021. Association for Computing Machinery.

\bibitem[\protect\citeauthoryear{Sun \bgroup \em et al.\egroup }{2022}]{enkt}
Jianwen Sun, Rui Zou, Ruxia Liang, Lu~Gao, Sannyuya Liu, Qing Li, Kai Zhang, and Lulu Jiang.
\newblock Ensemble knowledge tracing: Modeling interactions in learning process.
\newblock {\em Expert Systems with Applications}, 207:117680, 2022.

\bibitem[\protect\citeauthoryear{Tong \bgroup \em et al.\egroup }{2022}]{hgkt}
Hanshuang Tong, Zhen Wang, Yun Zhou, Shiwei Tong, Wenyuan Han, and Qi~Liu.
\newblock Introducing problem schema with hierarchical exercise graph for knowledge tracing.
\newblock In {\em Proceedings of the 45th International ACM SIGIR Conference on Research and Development in Information Retrieval}, SIGIR '22, page 405–415, New York, NY, USA, 2022. Association for Computing Machinery.

\bibitem[\protect\citeauthoryear{Touvron \bgroup \em et al.\egroup }{2023}]{llama}
Hugo Touvron, Thibaut Lavril, Gautier Izacard, Xavier Martinet, Marie-Anne Lachaux, Timoth{\'e}e Lacroix, Baptiste Rozi{\`e}re, Naman Goyal, Eric Hambro, Faisal Azhar, et~al.
\newblock Llama: Open and efficient foundation language models.
\newblock {\em arXiv preprint arXiv:2302.13971}, 2023.

\bibitem[\protect\citeauthoryear{Veličković \bgroup \em et al.\egroup }{2018}]{gat}
Petar Veličković, Guillem Cucurull, Arantxa Casanova, Adriana Romero, Pietro Liò, and Yoshua Bengio.
\newblock Graph attention networks.
\newblock In {\em \rc{International Conference on Learning Representations}}, 2018.

\bibitem[\protect\citeauthoryear{Wang \bgroup \em et al.\egroup }{2019}]{dgl}
Minjie Wang, Da~Zheng, Zihao Ye, Quan Gan, Mufei Li, Xiang Song, Jinjing Zhou, Chao Ma, Lingfan Yu, Yu~Gai, Tianjun Xiao, Tong He, George Karypis, Jinyang Li, and Zheng Zhang.
\newblock Deep graph library: A graph-centric, highly-performant package for graph neural networks, 2019.

\bibitem[\protect\citeauthoryear{Wang \bgroup \em et al.\egroup }{2021}]{egat}
Ziming Wang, Jun Chen, and Haopeng Chen.
\newblock Egat: Edge-featured graph attention network.
\newblock In {\em International Conference on Artificial Neural Networks}, pages 253--264. Springer, 2021.

\bibitem[\protect\citeauthoryear{Wu and Ling}{2023}]{self-sup-graph}
Tangjie Wu and Qiang Ling.
\newblock Self-supervised heterogeneous hypergraph network for knowledge tracing.
\newblock {\em Information Sciences}, 624:200--216, 2023.

\bibitem[\protect\citeauthoryear{Yang \bgroup \em et al.\egroup }{2020}]{gikt}
Yang Yang, Jian Shen, Yanru Qu, Yunfei Liu, Kerong Wang, Yaoming Zhu, Weinan Zhang, and Yong Yu.
\newblock Gikt: a graph-based interaction model for knowledge tracing.
\newblock In {\em Joint European Conference on Machine Learning and Knowledge Discovery in Databases}, pages 299--315. Springer, 2020.

\bibitem[\protect\citeauthoryear{Zhang and Chen}{2020}]{igmc}
Muhan Zhang and Yixin Chen.
\newblock Inductive matrix completion based on graph neural networks.
\newblock In {\em \rc{International Conference on Learning Representations}}, 2020.

\bibitem[\protect\citeauthoryear{Zhang \bgroup \em et al.\egroup }{2017}]{dkvmn}
Jiani Zhang, Xingjian Shi, Irwin King, and Dit-Yan Yeung.
\newblock Dynamic key-value memory networks for knowledge tracing.
\newblock In {\em Proceedings of the 26th international conference on World Wide Web}, pages 765--774, 2017.

\bibitem[\protect\citeauthoryear{Zhu \bgroup \em et al.\egroup }{2022}]{social_hypergraph}
Zirui Zhu, Chen Gao, Xu~Chen, Nian Li, Depeng Jin, and Yong Li.
\newblock \rc{Inhomogeneous Social Recommendation with Hypergraph Convolutional Networks}.
\newblock {\em 2022 IEEE 38th International Conference on Data Engineering (ICDE)}, 2022.

\end{thebibliography}
